\documentclass[sigconf]{acmart}
\usepackage{booktabs} 

\usepackage{subfigure}
\usepackage{bm}
\usepackage{xspace}
\usepackage{multirow}
\usepackage{footmisc}
\usepackage{amssymb}
\usepackage{array}
\usepackage{graphicx}
\usepackage{clrscode}
\usepackage{balance}
\usepackage{subfigure}
\usepackage{multirow}
\usepackage{multicol}
\usepackage{float}
\usepackage{color}
\usepackage{xcolor}
\usepackage{amsopn}
\usepackage{mathrsfs}
\usepackage{mathtools}
\usepackage{amsmath}
\usepackage{booktabs}
\usepackage{arydshln}
\usepackage{hyperref}
\usepackage{blkarray}
\usepackage{enumerate}
\usepackage{courier}
\usepackage{mathrsfs}
\usepackage{rotating}
\usepackage{bm}
\usepackage{subfigure}
\usepackage{array}
\usepackage{ragged2e}
\usepackage{amsthm}
\usepackage{verbatimbox}
\usepackage[ruled,linesnumbered]{algorithm2e}

\newcommand{\paratitle}[1]{\vspace{1ex}\noindent\textbf{#1}}
\newcommand{\ie}{\emph{i.e.,}\xspace}

\newcommand{\eg}{\emph{e.g.,}\xspace}

\newcommand{\ignore}[1]{}
\newcommand{\changed}[1]{\textcolor{blue}{#1}}

\copyrightyear{2020}
\acmYear{2020}
\setcopyright{acmcopyright}\acmConference[CIKM '20]{Proceedings of the 29th ACM
	International Conference on Information and Knowledge Management}{October
	19--23, 2020}{Virtual Event, Ireland}
\acmBooktitle{Proceedings of the 29th ACM International Conference on Information
	and Knowledge Management (CIKM '20), October 19--23, 2020, Virtual Event, Ireland}
\acmPrice{15.00}
\acmDOI{10.1145/3340531.3411929}
\acmISBN{978-1-4503-6859-9/20/10}

\AtBeginDocument{%
	\providecommand\BibTeX{{%
			\normalfont B\kern-0.5em{\scshape i\kern-0.25em b}\kern-0.8em\TeX}}}

\begin{document}
\fancyhead{}
\title{Learning to Match Jobs with Resumes from Sparse Interaction Data using Multi-View Co-Teaching Network}

\author{Shuqing Bian$^{1}$, Xu Chen$^{4}$, Wayne Xin Zhao$^{2,3*}$, Kun Zhou$^{1}$, }\thanks{$^*$Corresponding author.} 
\author{Yupeng Hou$^{2}$, Yang Song$^{5}$, Tao Zhang$^{5}$ and Ji-Rong Wen$^{2,3}$}

\affiliation{%
	\institution{$^1$School of Information, Renmin University of China}
	\institution{$^2$Gaoling School of Artificial Intelligence, Renmin University of China}
	\institution{$^3$Beijing Key Laboratory of Big Data Management and Analysis Methods}
	\institution{$^4$Department of Computer Science, University College London}
	\institution{$^5$BOSS Zhipin NLP Center}
}
\affiliation{%
	\institution{ \{bianshuqing, houyupeng, jrwen\}@ruc.edu.cn, xu.chen@ucl.ac.uk, batmanfly@gmail.com,}
	\institution{francis\_kun\_zhou@163.com, \{songyang, kylen.zhang\}@kanzhun.com}
}

\begin{abstract}
With the ever-increasing growth of online recruitment data, \emph{job-resume matching} has become an important task to automatically match jobs with suitable resumes.
This task is typically casted as a supervised text matching problem. Supervised learning is powerful when the labeled data is sufficient. However, on online recruitment platforms, job-resume interaction data is sparse and noisy, which affects the performance of job-resume match algorithms.

\ignore{This task is typically casted as a supervised text matching problem.While supervised learning is powerful when the labeled data is sufficient and clean, the job-resume interaction in practice is usually sparse and noisy, which brings difficulties to effective text representation learning.}

To alleviate these problems, in this paper, we propose a novel multi-view co-teaching network from sparse interaction data for job-resume matching. 
Our network consists of two major components, namely text-based matching model and relation-based matching model. 
The two parts capture semantic compatibility in two different views, and complement each other. 
In order to address the challenges from sparse and noisy data, we design two specific  strategies 
to combine the two components.
First, two components share the learned parameters or representations, so that the original representations of each component can be enhanced. More importantly, we adopt a co-teaching mechanism to reduce the influence of noise in training data. The core idea is to let the two components help each other by selecting more reliable training instances. The two strategies focus on \emph{representation enhancement} and \emph{data enhancement}, respectively. 
Compared with pure text-based matching models, the proposed approach is able to learn better data representations from limited or even sparse interaction data, which is more resistible to noise in training data.
 Experiment results have demonstrated that our model is able to outperform state-of-the-art methods for job-resume matching. 
\end{abstract}


\maketitle

{\fontsize{8pt}{8pt} \selectfont
	\textbf{ACM Reference Format:}\\
Shuqing Bian, Xu Chen, Wayne Xin Zhao, Kun Zhou, and Yupeng Hou, Yang Song, Tao Zhang and Ji-Rong Wen. 2020. Learning to Match Jobs with Resumes from Sparse Interaction Data using Multi-View Co-Teaching Network. In \textit{Proceedings of the 29th ACM International Conference on Information and Knowledge Management (CIKM '20), October 19--23, 2020, Virtual Event, Ireland.} ACM, New York, NY, USA, 10 pages. https://doi.org/10.1145/3340531.3411929 }

\section{Introduction}
Nowadays, online recruitment platforms play an increasingly important role in connecting job seekers with employers. 
According to a report from LinkedIn\footnote{https://www.businessofapps.com/data/linkedin-statistics/}, there were 660 million users and 20 million job listings on LinkedIn from about over 200 countries and territories all over the world as of late November 2019.
With the huge amount of online recruitment data, it has become an essential task to automatically match jobs with suitable candidates, called \emph{job-resume matching}, which aims to accelerate the recruitment process. 

In online recruitment systems,  employers publish the \emph{job postings} (referred to \emph{jobs} for short) describing what qualifications areas are essential to satisfactory performance in a position, and job seekers upload the \emph{resumes} stating their skills, experience, and attributes. 
Basically, the two kinds of documents  are mainly written in natural languages. 
Hence, a typical approach is to cast the job-resume matching task as a supervised text matching problem~\cite{QinZXZJCX18,ShenZZXMX18}.
In such a setting, these methods  rely on a set of matched or unmatched job-resume pairs~(termed as \emph{job-resume interaction}) as training data. Apparently, the amount and quality of available job-resume interaction data directly affect the performance of job-resume matching algorithms. 

However,  on online recruitment platforms, job-resume interaction data is extremely sparse, and likely to contain noise. As shown in the report~\cite{RamanathIPHGOWK18}, on average, a job posting only has about three candidates for final interview. 
The major reason is that employers will carefully evaluate 
whether a job seeker is suitable for this job before sending an interview acceptation.
While, other kinds of interactive  behaviors (\eg chatting online and requesting profiles) are not accurate to reflect the final status for job-resume matching, so that it is not reliable to utilize interaction data as training data.  Especially, negative feedbacks~(\ie explicit rejection) are more difficult to obtain in practice. 
Many employers may not actively send an explicit rejection notification to unqualified candidates. 
 In order to obtain sufficient negative samples, a simple way is to sample random pairs without acceptation status~\cite{XuFHZ15}, called \emph{negative sampling}.  Such a way is problematic since it will incorporate noisy labeled data. For example, although online interaction has been finished without an explicit success,  an employer and job seeker may actually schedule the interview via phones in an offline way.

\begin{figure}[htbp]
   \centering
	\includegraphics[width=0.5\textwidth]{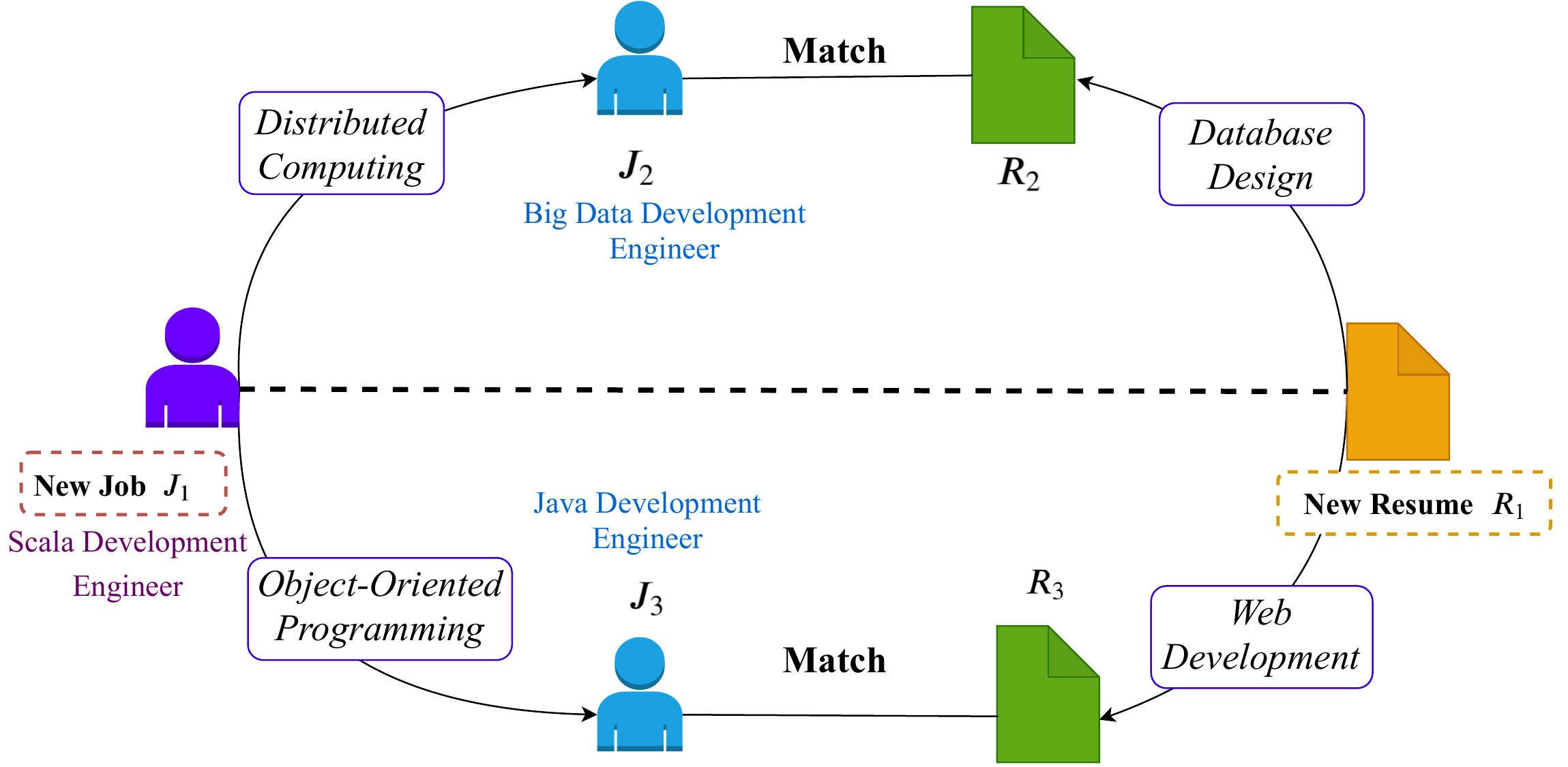}
	\caption{A motivating example for relation-based match. Here, we present three jobs $\{j_1, j_2, j_3\}$ and three resumes $\{r_1, r_2, r_3\}$, where $\langle j_2, r_2 \rangle$ and $\langle j_3, r_3 \rangle$ are matched pairs in training data. }
	\label{fig:intro}
\end{figure}

To address the above issues, we consider capturing multi-view  data signals for alleviating the sparsity of explicit interaction data.
Inspired by  collaborative filtering algorithms in recommender systems~\cite{SarwarKKR01,JiangFXCM19}, our idea is to mine underlying correlations among jobs and resumes for enhancing the representations by aggregating  useful evidence from similar jobs or resumes.  
Since  job-resume interaction data itself is sparse, we do not directly learn the correlation characteristics from such interaction data. 
Instead, we notice that  job postings or resumes are usually written in a skill-oriented way and associated with  a specified  position. Containing the same skill keywords or position labels  is an important indicator for the correlation between two documents. 
Figure~\ref{fig:intro}  has presented a motivating example for our idea. As we can see, job $j_1$ is a minor position with few historical matched cases.  It is relatively difficult to directly match it with suitable candidates. While,  $j_1$ shares some same skill requirements with two other  jobs of $j_2$ and $j_3$.
Interestingly,  $j_2$ and $j_3$ have historical matched resumes $r_2$ and $r_3$, respectively. 
Furthermore, we can indeed find that resume $r_1$ contains  composite  skills that are contained in $r_2$ and $r_3$. Although these skills might not be exactly the same as those required by $j_1$, it is likely that $r_1$ is a good candidate for $j_1$.
In this case, we can see that relation-based semantic signal is potentially useful to improve a pure text-based matching model. 
Hence, we aim to develop a multi-view learning approach that is able to effectively integrate both kinds of data signals for deriving a more capable, robust matching model. 

To this end, in this paper, we propose a novel multi-view co-teaching network from sparse interaction data for job-resume matching.
Our network consists of two major components, namely text-based matching model and relation-based matching model. The two parts capture semantic compatibility in two different views, and are designed to complement each other.   
The text-based matching component is implemented by a hierarchical self-attention text encoder, using the 
Transformer architecture~\cite{VaswaniSPUJGKP17} and the pre-trained BERT~\cite{DevlinCLT19} model.
In this way, we can obtain the corresponding text representations of jobs and resumes. 
For the relation-based matching component, we first construct the job-resume relation graph, in which jobs or resumes are considered as nodes and their relation-specific connections are considered as links. 
Furthermore, we develop a relational graph neural network for learning node representations based on the job-resume relation graph.
Having both components, the key point is how to integrate them into a unified approach. We design two strategies for combining the merits of the two models. First, we let the two parts share the learned parameters or representations, so that the original representations of each part can be enhanced. 
More importantly, we adopt a co-teaching mechanism to reduce the influence of noise  in training data. 
The core idea is to  let the two components help each other by selecting more reliable training instances.
The two strategies focus on \emph{representation enhancement} and \emph{data enhancement}, respectively.
Compared with pure text-based matching models, the proposed approach is able to learn better representations from limited or even sparse interaction data, which is more  resistible to noise in training data.  

To evaluate the effectiveness of our model, we conduct extensive experiments on three real-world datasets. 
Experimental results demonstrate that our model outperforms several state-of-the-art methods for job-resume matching. 
To the best of our knowledge, it is the first time that both text- and relation-based matching models are integrated into a unified approach for the job-resume matching task. 
Our approach specially considers the sparsity issue of training data and is also resistible to noisy data.

\section{Related Work}
\hyphenpenalty=8000
\tolerance=2000
This paper aims to learn an effective job-resume matching model with noisy interaction data.
Thus we review the related work in the fields of job-resume matching and learning with noisy labels respectively.

\paratitle{Job-resume matching}. 
Matching jobs and resumes stands at the core of a recruitment platform. As an important task in recruitment data mining~\cite{ShalabyAKPAQZ17,KenthapadiLV17}, person-job fit has been extensively studied in the literature.
Early methods cast this problem as a recommendation task~\cite{lu2013recommender,diaby2013toward}, and the matching capability is obtained based on the collaborative filtering assumption.
To alleviate this problem, recent research mostly focused on text matching technology, where the basic problem is how to represent the document.
Around this problem, 
Shen \emph{et al.}~\cite{ShenZZXMX18} proposed to encode the job and resume based on CNN.
Qin \emph{et al.}~\cite{QinZXZJCX18} leveraged RNN and BiLSTM to capture textual sequential information for more accurate semantic representation.
In order to discriminate different sentence importances so as to achieve better encoding accuracy, Yan \emph{et al.}~\cite{YanLSZZ019} proposed a profiling memory module to learn the latent preference representation by interacting with both the job and resume sides. Bian \emph{et al.}~\cite{BianZSZW19} used hierarchical attention-based RNN to match the jobs and resumes.
In addition, Luo \emph{et al.}~\cite{LuoZWZ19} studied the effectiveness of adversarial training for the job-resume matching problem.
Unlike these models, which focus on either the relations or the semantics of the jobs and resumes, we combine both of these information by a unified framework to alleviate the data sparsity and noisy problems.

\paratitle{Learning with noisy labels}. 
Basically, learning with noisy labels aims to solve the problem when some of the training data is unreliable in supervised learning.
Early methods, such as the curriculum learning ~\cite{BengioLCW09} and self-paced learning ~\cite{KumarPK10}, alleviate this problem by reordering the training samples.
Easy instances will be learned before the harder ones. 
Following this idea, many variants have been proposed, such as the methods based on deep reinforcement learning ~\cite{FanTQ0L18}, MentorNet ~\cite{JiangZLLF18} and UBD~\cite{MalachS17}.
In parallel to the curriculum learning, several studies leverage different weighting mechanisms to lower the impact of the noisy instances ~\cite{WuTXFQLL18}.

Our paper is inspired from a recent work called co-teaching network (CTN)~\cite{HanYYNXHTS18}. The co-teaching network adopts a \emph{learning to teach} strategy for dealing with noise~\cite{abs-04413} and unlabeled data~\cite{YangLZLSCGHJL20}.
However, there are several significant differences with our work. First, CTN identifies the noisy samples within the same view, while our model leverages different information (\ie relation and text) to examine noisy instances.
In addition, we design a ``re-weighting'' mechanism tailored for one-class classification problem, and apply it into the job-resume matching task.

\section{PROBLEM DEFINITION}
Suppose that we have a  set  of jobs $\mathcal{J}=\left\{j_1, j_2, \cdots, j_{n}\right\}$ and a  set of resumes $\mathcal{R}=\left\{r_{1}, r_{2}, \cdots ,r_{m}\right\}$, where $n$ and $m$ are the total numbers of jobs and resumes, respectively. 
Each job or resume is represented by a text document describing the position's requirements or the applicant's skills, respectively.
We are also given an observed (training) matching set $\mathcal{Y}=\left\{ \langle j, r, y_{j,r} \rangle | j \in \mathcal{J},  r \in \mathcal{R} \right\}$, where $y_{j,r}$ is a binary label indicating the final match result between job $j$ and resume $r$: \emph{accept}~(\textsc{Yes}) or \emph{reject}~(\textsc{No}).
Based on the above notations, our task is to learn a predictive function $f(j',r')$ based on the matching set $\mathcal{Y}$, so that it can accurately estimate the matching degree for an unseen (testing) job-resume pair $\langle j', r' \rangle$.
In practice, the interaction data for the job-resume matching task is usually extremely sparse (\emph{i.e.}, $n\times m \gg |\mathcal{Y}|$), 
and the training data may be also noisy (\eg the randomly sampled negative instances). 
Previous methods mainly focus on learning effective text representations, and cast the task as a supervised text matching task. Especially, they seldom consider the influence of the quality of training data on the model performance. In what follows, we present our multi-view co-teaching network for addressing these issues, where we characterize the matching patterns from different views, and leverage their complementary features to improve the training instances.

\section{THE PROPOSED APPROACH}
\begin{figure}[htbp]
	\centering
	\includegraphics[width=0.45\textwidth]{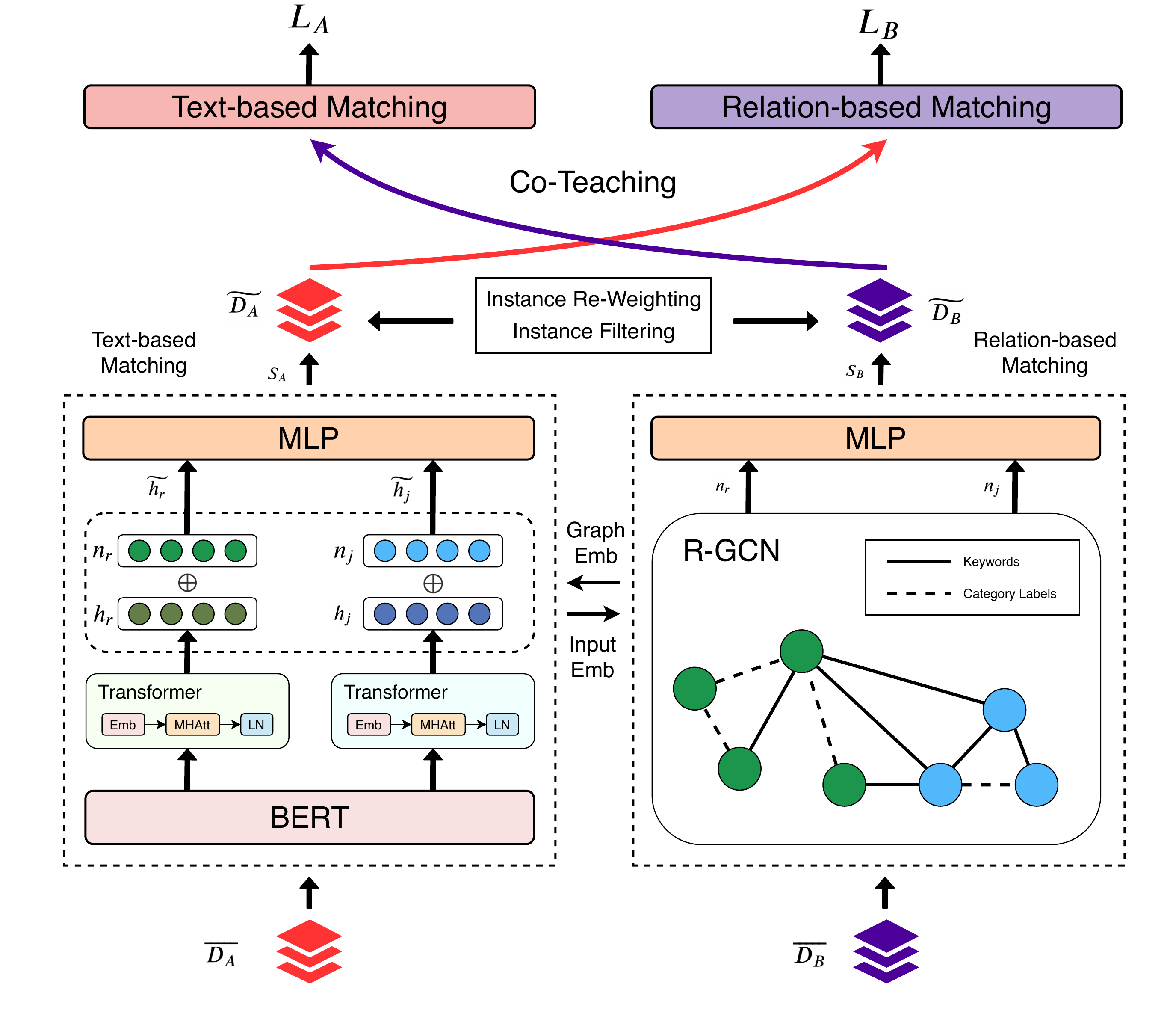}
	\caption{The overall architecture of our proposed approach.}
	\label{fig:model}
\end{figure}

In this section, we introduce the  proposed approach for the job-resume matching task in detail.
The overall framework is presented in Figure~\ref{fig:model}.
On one hand, jobs and resumes are described in text documents. 
We adopt a hierarchical self-attention text encoder for capturing text semantics,  called \emph{text-based matching model}.
On the other hand, we construct a relation graph, regarding the jobs and resumes as nodes and their underlying correlations as links. 
We cast this task as link prediction and develop a graph neural network based model, called \emph{relation-based matching model}. 
As motivated in Section 1, the two models have their own merits, so we further integrate them into a unified multi-view co-teaching network. 
Figure~\ref{fig:model} presents the overall architecture of our proposed approach. Next, we describe each part in detail.

\subsection{Text-based  Matching Component}
For text-based matching approaches, they seek to represent job text and resume text  in a suitable way, and then construct the matching model based on semantic similarity.
A key difficulty lies in how to effectively represent the job and resume documents. 

Many text representation models have been explored in this task, including  LSTM~\cite{ZhouQZXBX16} and CNN~\cite{Kim14}. More recently,  self-attention mechanisms~(\eg Transformer~\cite{VaswaniSPUJGKP17}) and its extensions on pre-trained model~(\eg BERT~\cite{DevlinCLT19}) have made great progress in various natural language processing tasks.
However, BERT usually has a length limit on the input text, \eg 512 words, which prevents the accurate modeling of long documents. Besides, we believe that the sentence boundary should be useful signal to text representations, \eg a sentence corresponds to a skill requirement.
Based on above considerations, we develop a hierarchal self-attention text representation model for developing the semantic matching model, in which a BERT-based encoder is first adopted to represent sentences, and then a Transformer-based encoder is used to represent the overall document based on learned sentence embeddings.

\subsubsection{BERT-based Sentence Encoder} The first layer of our model is a sentence encoder 
implemented by a standard BERT model, which is a bidirectional Transformer with multiple layers. Given a job or resume sentence, 
a special token ``\emph{CLS}'' has been inserted into the beginning of the sentence. 
For each token in the input sentence, two kinds of embeddings are considered as input, including  token embeddings indicate the meaning of each token,  and position embeddings indicate the position of each token within the text sequence. These two embeddings are summed to a single input vector and fed to the BERT encoder. 
The learned representation for the ``\emph{CLS}'' symbol is treated as the sentence representation.
\subsubsection{Hierarchical Transformer Encoder}
The document encoder is developed on top of the BERT-based sentence encoder. 
Given a job or resume, it takes as input the sentence embeddings and outputs the overall document representation. Using a hierarchical architecture, our encoder is able to model very long documents, and also keep the sentence boundaries. Formally, the update formulas for our document encoder are given as follows:

\begin{eqnarray}
\tilde{\bm{h}}^{(l)}_{r}&=&\mathrm{LN}\left(\bm{h}^{(l-1)}_{r}+\mathrm{MHAtt}\left(\bm{h}^{(l-1)}_{r}\right)\right),\\
 \tilde{\bm{h}}^{(l)}_{j}&=&\mathrm{LN}\left(\bm{h}^{(l-1)}_{j}+\mathrm{MHAtt}\left(\bm{h}^{(l-1)}_{j}\right)\right),\\
\bm{h}^{(l)}_{r}&=&\mathrm{LN}\left(\tilde{\bm{h}}^{(l)}_{r}+\mathrm{FFN}\left(\tilde{\bm{h}}^{(l)}_{r}\right)\right),\\
 \bm{h}^{(l)}_{j}&=&\mathrm{LN}\left(\tilde{\bm{h}}^{(l)}_{j}+\mathrm{FFN}\left(\tilde{\bm{h}}^{(l)}_{j}\right)\right),
\end{eqnarray}
where $j$ denotes a job document, $r$ denotes a resume document,  ${\bm{h}}^{(l)}_{r}$ and ${\bm{h}}^{(l)}_{j}$ are the $l$-th layer input resume and job vectors,  \emph{LN} is the layer normalization operation, and \emph{MHAtt} is the multi-head attention operation. 



Let $L$ denote the number of layers in the Transformer network. 
The final output layer is a sigmoid classifier, defined as
\begin{eqnarray}\label{eq-match1}
\hat{y}_{j,r}=\sigma\left(\bm{W}_{1}\left[ \bm{h}_{j}^{(L)}; \bm{h}_{r}^{(L)}\right]+b_{1}\right),
\end{eqnarray}
where   $\bm{h}_{j}^{(L)}$ and $\bm{h}_{r}^{(L)}$ are the  representations at the final layer (\ie the $L$-th layer) for job document $j$ and resume document $r$ respectively,  $\bm{W}_{1}$ is a parameter matrix for transforming the concatenated job and resume document representations, $b_1$ is a bias, and $\hat{y}_{j,r} \in (0,1)$ indicates the matching degree between job $j$ and resume $r$.

\subsection{Relation-based Matching Component}

In the above model,  jobs and resumes are matched according to their text content. The matching results are derived from their semantic compatibility. 
Here, we take a new perspective to study the job-resume matching task. Intuitively, there exist implicit correlations among jobs and resumes. For example, two similar jobs will be attractive to the same applicant, and 
a company may have a few comparable resumes for a job position. 
Since explicit matching interaction (no matter \emph{success} or \emph{failure}) is sparse, mining underlying implicit correlation will be useful to extract additional signals to complement semantic compatibility.
For this purpose, we first construct a job-resume relation graph for capturing implicit correlations, and then  develop a matching model using relational graph neural networks. 

\subsubsection{Construction of the Job-Resume Relation Graph}
Before presenting the specific construction algorithms, we first formally define the job-resume relation graph $\mathcal{G} = \{ \mathcal{V}, \mathcal{E} \}$, where $\mathcal{V}$ is the node set and the $\mathcal{E}$ is the link set. For our task, the node set is the union between job set and resume set, \ie  $\mathcal{V}=\mathcal{J} \cup \mathcal{R}$, and the link set contains all the interaction links between two nodes on the graph, \ie $\mathcal{E}=\{ \langle v_1, v_2, t_{v_1, v_2} \rangle |  v_1, v_2 \in \mathcal{V} , t_{v_1, v_2} \in \mathcal{T} \}$, where $\mathcal{T}$ is the relation set and $t_{v_1, v_2}$ is the relation label for the link between $v_1$ and $v_2$.
Since we have two kinds of nodes, there are three possible types of links to consider in our task, \ie \emph{job-to-resume}, \emph{job-to-job} and \emph{resume-to-resume}.
We consider two kinds of data signals to create implicit links between two nodes, either   
\emph{category label} and \emph{keyword}.

\paratitle{Link Creation using Category Labels}. In an online recruitment platform, in order to organize the job and resumes, there is usually a job taxonomy, in which a higher (or lower) level indicates a more coarse-grained (or fine-grained) kind of job positions. We only consider the category labels at the bottom layer, which has  good discrimination capacity on specifying some kind of jobs, \eg \emph{JAVA software engineer} or \emph{accountant}.   
Here, we create links between two job nodes or two resume nodes if they share the same category label. 
A formed link is further attached with the corresponding category label as the relation label. 

\paratitle{Link Creation using Keywords}. In both job and resume documents, not all the keywords are equally informative for predicting the final match results. Some words may contain more important semantics~(\eg the description words for skill). Here, we would like to extract a list of keywords from both job and resume documents, and identify underlying correlation through such indicators. 
The overall algorithm can be described as follows.
First, we use the classic \emph{tf-idf} term weighting method to select a list of candidate words. 
Then, we construct a word co-occurrence graph by counting the frequency that two words co-occur in a job or resume document. Third, we  run a standard PageRank algorithm on the word co-occurrence graph and obtain the PageRank scores of each word. Finally, we sort the words according to their PageRank scores, and only keep the top $K$ words as the \emph{keywords}. 
When the text of two nodes~(either a job or a resume)  contains a same keyword, we create a link between them and attach the keyword as relation label. 

\subsubsection{Learning Job and Resume Representations}
Based on the job-resume relation graph, we further learn effective  representations for capturing the underlying semantics reflected by the graph for job-resume match.
Recently, graph neural networks have become one  popular class of models for learning node characteristics
from the graph-structured  data, \eg graph convolutional networks (GCN)~\cite{KipfW17}.
However,  traditional GCN mainly deals with homogeneous links, which cannot effectively characterize different types of links. Inspired by the recent progress made in knowledge graph completion~\cite{LinLSLZ15}, we adopt the 
Relational Graph Convolutional Network~(RGCN)~\cite{SchlichtkrullKB18} to model the relation-specific links. 
Based on GCN, the core difference of RGCN lies in the aggregation step, where we collect the incoming information from neighbors and perform the aggregation operation according to different types of relations. 

Formally, the node representation at the $l$-th layer is derived by its neighbors' embeddings at the previous layer as:
\begin{equation}\label{eq-GCN}
\bm{n}_{v}^{(l+1)}=\sigma\big(\sum_{t \in \mathcal{T}} \underbrace{\frac{1}{|\mathcal{V}_{v}^{t}|}\sum_{v' \in \mathcal{V}_{v}^{t}}\bm{W}_{t}^{(l)}\bm{n}_{v'}^{(l)}}_{\text{propagation \emph{w.r.t.} specific relations}}+\bm{W}^{(l)}_o \bm{n}_{v}^{(l)}\big),
\end{equation}
where $v$ is a node placeholder that can be a job node or a resume node, 
$\bm{n}_{v}^{(l)} \in \mathbb{R}^{D}$ is the representation of node $v$ at the $l$-th layer, 
$\mathcal{V}_{v}^{t}$ denotes the set of $v$'s neighbors under the relation type $t$, 
$\bm{W}_{t}^{(l)}$ is a parameter matrix specific to relation $t$, and  $\bm{W}^{(l)}_o$ is the parameter matrix specific to original node representation.

In this equation, each relation type $t$ is associated with a specific parameter matrix $\bm{W}_{t}^{(l)}$, so that node representations are able to incorporate relation-aware semantics and reduce the influence of irrelevant aspects.

Once we have derived the node representations, the matching score between a job and a resume can be defined in a similar way as Eq.~\ref{eq-match1}:
\begin{eqnarray}\label{eq-match2}
\hat{y}_{j,r}=\sigma\left(\bm{W}_{2}\left[ \bm{n}_{j}^{(L')}; \bm{n}_{r}^{(L')}\right]+b_{2}\right),
\end{eqnarray}
where is the predicted  $\bm{n}_{j}^{(L')}$ and $\bm{n}_{r}^{(L')}$ are the representations at the final layer for job document $j$ and resume document $r$ respectively, and $\bm{W}_{2}$ and $b_2$ are the parameter matrix or vector.

\subsection{Multi-View Co-Teaching Network}
Previously, we have described two individual components for job-resume matching from different perspectives, namely text- and relation-based views. Each of the two models has its own merits on our task, and we further study how to integrate them into a unified approach. 

We design two integration strategies and develop a multi-view co-teaching network.
 First, we share the learned information or parameters for enhancing the original representations of each components. Second, as motivated in Section 1, we focus on reducing the influence of noise from training data, especially for negative samples. 
We borrow the idea of co-teaching method~\cite{HanYYNXHTS18} introduced in machine learning, and let the two components help each other by selecting more reliable training instances. 
Next, we present the details of the two strategies.

\subsubsection{Representation Enhancement} Recall that for a job $j$  we have learned two kinds of different representations for both jobs and resumes, namely the text-based representation $\bm{h}_j$ and relation-based representation $\bm{n}_j$. Similarly for a resume $r$, we have the representations of $\bm{h}_r$ and $\bm{n}_r$, correspondingly. 

To enhance the semantic matching model, we concatenate the original text-based representation with the relation-based representation as:
\begin{eqnarray}
\widetilde{\bm{h}}_j &=& \bm{h}_j \oplus \bm{n}_j,\\
\widetilde{\bm{h}}_r &=& \bm{h}_r \oplus \bm{n}_r,
\end{eqnarray}
where ``$\oplus$'' is the vector concatenation operation. Furthermore, we feed  the enhanced representations into Eq.~\ref{eq-match1} for achieving a better prediction result. For the relation-based matching model, we do not directly adopt the above concatenation method. Instead,  the initialization of the RGCN model is particularly important to the final performance. Hence, we utilize the learned representations from semantic matching model for initializing node states:
\begin{eqnarray}
\bm{n}_{j}^{(0)}  &=& \bm{h}_j,\\
\bm{n}_{r}^{(0)}  &=& \bm{h}_r.
\end{eqnarray}

\ignore{For taking the best of both worlds, we can ensemble them by fusing their output or sharing some intermediate layers.
However, in our problem, such simple combining methods can be problematic, because the sampled negative instances can be noisy, and the ensemble methods have no mechanism to prohibit the noisy information from being recursively transfered from one model to its peer, which may greatly degrade the model performance.
}

\subsubsection{Co-Teaching for Data Enhancement} In this part, we study the second integration strategy for enhancing the quality of training data. Our basic assumption is that true samples usually receive  similar predictions under different model views, while the noisy ones are  not easy to cheat all the models.
In the co-teaching framework, our two components can be considered as two peer learners.
The samples for training one learner will be firstly examined by the other one, and only the instances labeled as ``high quality'' will be kept in the training phrase.
Since the two learners have  very different views to model the data characteristics, it is expected that they can help each other to select ``high quality'' training samples, and thus improve the final performance.

\paratitle{Overall Algorithm}. The overall co-teaching process is presented in Algorithm~\ref{alg:co-teaching}.
Formally, let  $A$ and $B$ denote text-based and relation-based matching models, respectively.
For each batch update, a batch dataset $\overline{\mathcal{D}}$ is randomly split into two equal-sized subsets namely  $\overline{\mathcal{D}}_{A}$ and $\overline{\mathcal{D}}_{B}$  (\emph{line} 4).
Instead of directly feeding them into model $A$ and $B$, $\overline{\mathcal{D}}_{A}$ and $\overline{\mathcal{D}}_{B}$ are firstly ``examined'' by their peer learner to filter noisy samples (\emph{line} 5-6).
Finally, based on the derived new datasets $\widetilde{\mathcal{D}}_{A}$ and $\widetilde{\mathcal{D}}_{B}$, the parameters of $A$ and $B$ will be updated using stochastic gradient decent (SGD) (\emph{line} 7-8).
As we can see, the key point of this algorithm lies in peer examination (\emph{line} 5-6).
Next, we give two implementations for peer examination.  

\ignore{The overall training process can be seen in Algorithm~\ref{alg:co-teaching}.
In specific, each training batch $\bar{D}$ is split equally into two sub-batches $\bar{D}_{A}$ and $\bar{D}_{B}$ (line 3).
Instead of feeding into model $A$ and $B$, $\bar{D}_{A}$ and $\bar{D}_{B}$ are firstly ``examined'' by their peers to filter noisy samples (line 4-5).
Based on the derived sub-batches and the loss functions, the parameters of model $A$ and $B$ are learned based on stochastic gradient decent (SGD) (line 6-7).
}
\begin{algorithm}[t]
	\caption{The proposed co-teaching algorithm.}
	\label{alg:co-teaching}
	\LinesNumbered
	\KwIn{\\
		The set of training data, $\mathcal{D}$;\\
		Model parameters $\theta_{A}$, $\theta_{B}$; \\
		Epoch $E$ , iteration $N$, learning rate $\eta$;
	}
	\KwOut{$\theta_{A}$, $\theta_{B}$;}
	\For{$b=1,2, \ldots, N$}{	
		\textbf{Shuffle} training set $\mathcal{D}$;\\
		\For{$e=1,2, \dots, E$}{
			\textbf{Fetch} a batch of training data $\overline{D}$;\\
			\textbf{Learning} $(\widetilde{\mathcal{D}}_{B}, L_{B})$ from model $A$ and $\overline{\mathcal{D}}_{B}$;\\
			\textbf{Learning} $(\widetilde{\mathcal{D}}_{A}, L_{A})$ from model $B$ and $\overline{\mathcal{D}}_{A}$;\\
			\textbf{Update} $\theta_{A}=\theta_{A}-\eta \nabla L_{A}(\widetilde{\mathcal{D}}_{A})$;\\
			\textbf{Update} $\theta_{B}=\theta_{B}-\eta \nabla L_{B}(\widetilde{\mathcal{D}}_{B})$;
		}
	}
\end{algorithm}

\paratitle{Training with Instance Re-Weighting}. The first training method is to re-weight the training instances. Given a model, its peer model aims to increase the weight of ``high-quality'' samples and decrease the weight of unreliable samples from its perspective. In order to train model $B$, we assume that  $K$  instances $\overline{\mathcal{D}}_{B}=\{ \langle  j_i, r_i, y_i \rangle \}_{i=1}^K$ are generated during a batch update in the training process. 
We let model $A$ assign a weight to each instance in $\overline{\mathcal{D}}_{B}$, denoted by $w_i$. 
The key idea is to penalize the instance according to the disagreement degree between the given information and  $A$'s prediction: 
\begin{equation}\label{eq-weight}
w_i^A = 1 - s_A(j_i, r_i, y_i),
\end{equation}
where $s_A(j_i, r_i, y_i)$ is the confidence score by model $A$ to predict the label of $y_i$ for the pair $\langle j_i, r_i \rangle$. After instance re-weighting, we can generate a new batch training dataset by augmenting the original instances with the weights, 
$\widetilde{\mathcal{D}}_{B}=\{ \langle  j_i, r_i, y_i, w_i^A \rangle \}_{i=1}^K$. Then, we can rewrite the loss function in a batch to train model $B$ as:

\begin{eqnarray}\label{eq-lossLB}
L_{B}(\overline{\mathcal{D}}_{B})=\sum_{i=1}^{K} w_i^A \cdot g\left(y_i, y_i^B\right),
\end{eqnarray}
where $g\left(y_i, y_i^B\right)$ is the loss for $B$'s prediction $y_i^B$. 
As mentioned before, the noisy information is mainly from negative sampling. Hence, we fix the weights for positive instances and true negative instances as 1 in the learning process, and the instances obtained by sampling from the dataset are weighted according to Eq.~\ref{eq-weight}.

\paratitle{Training with Instance Filtering}. 
Besides re-weighting different samples, we can also directly drop the instances which are not ``good enough''. Intuitively, if an instance can lead to a small loss value by a model, then it is far from the decision boundary, and is more likely to be a reliable sample instead of noises.
This idea is modeled by the following formula:
\begin{equation}\label{eq-filter}
\widetilde{\mathcal{D}}_{B} \leftarrow \operatorname{argmin}_{ \widetilde{\mathcal{D}}_{B} \subset \overline{\mathcal{D}}_{B}, \left|\widetilde{\mathcal{D}}_{B}\right|=\delta |\overline{\mathcal{D}}_{B} |}  L_{A}\left(\widetilde{\mathcal{D}}_{B}\right),
\end{equation}
where $L_{A}\left(\widetilde{\mathcal{D}}_{B}\right)$ denotes the accumulative loss of $A$ on the given dataset, and  $\delta$ is a hyper-parameter defining the held-out rate. Here, we select a subset from  the original data $\overline{\mathcal{D}}_{B}$ that is able to minimize the loss from the peer model.

\ignore{So, in our framework, the batches $\overline{{\mathcal{D}}}_{A}$ and $\overline{{\mathcal{D}}}_{B}$ are filtered by selecting the samples with smallest loss values as:
\begin{eqnarray}
\begin{aligned}
&\tilde{\mathcal{D}}_{B}=\operatorname{argmin}_{\left|\tilde{\mathcal{D}}_{B}\right|=\delta |\overline{\mathcal{D}}_{B} |, \tilde{\mathcal{D}}_{B} \subset \overline{\mathcal{D}}_{B}}  \mathcal{J}_{A}\left(\tilde{\mathcal{D}}_{B}\right)\\
&\tilde{\mathcal{D}}_{A}=\operatorname{argmin}_{\left|\tilde{\mathcal{D}}_{A}\right|=\delta |\overline{\mathcal{D}}_{A} |, \tilde{\mathcal{D}}_{A} \subset \overline{\mathcal{D}}_{A}}  \mathcal{J}_{B}\left(\tilde{\mathcal{D}}_{A}\right)
\end{aligned}
\end{eqnarray}
where | $\cdot$ | measures the size of a set, $\mathcal{J}_{A}\left(\tilde{\mathcal{D}}_{B}\right)$ and $\mathcal{J}_{B}\left(\tilde{\mathcal{D}}_{A}\right)$ stand for accumulation of loss on the corresponding data sets, and $\delta$ is a hyper-parameter.
}

In the above, we only discuss the case of updating $B$ with peer model $A$. It is similar to update model $A$ with peer model $B$.
Both instance re-weighting and filtering methods aim to select more reliable samples for model learning. They achieve this purpose with different approaches.  
Instance re-weighting is a ``soft'' method, where all the instances are remained, but with different training importances. As a comparison, 
instance filtering is a ``hard'' method, where some samples are directly dropped. 
One can also combine these two methods by filtering the samples before re-weighting them.

\paratitle{Complexity Analysis}. 
Compared with traditional supervised job-resume matching methods, the increased training complexity for instance re-weighting depends on the loss computation in Eq.~\ref{eq-lossLB}.
Suppose the complexities for computing the loss (or weight) of a sample by model $A$ and $B$ are $C_A$ and $C_B$, respectively.
Then the total additional complexity is $\mathcal{O}(NM(C_A+C_B))$ for instance re-weighting, where $N$ and $M$ are the numbers of training epochs and instances, respectively.
For the method of instance filtering, we need to compute the sampling loss and  
rank them for selecting a candidate set.
Since the ranking operation takes a cost of $\mathcal{O}(NM\log M)$, the total additional complexity for this method is $\mathcal{O}(NM(C_A+C_B+\log M))$.
\section{EXPERIMENTS}
In this section, we conduct extensive experiments to verify effectiveness of our model. We mainly address the following research questions:

\ignore{
\begin{itemize}
	\item \textbf{RQ1.} Whether our method can outperform the state-of-the-art job-resume matching models? \\
	\item \textbf{RQ2.} What are the effects of different components in our model? \\
	\item \textbf{RQ3.} How  the performance of our model varies with different parameters or settings? \\
	\item \textbf{RQ4.} What is the effect of our co-teaching mechanism from the qualitative perspective?\\
\end{itemize}
}

$\bullet$ \textbf{RQ1.} Whether our method can outperform the state-of-the-art job-resume matching models? 

$\bullet$ \textbf{RQ2.} What are the effects of different components in our model? 

$\bullet$ \textbf{RQ3.} How does the performance the performance of our model vary with different parameters or settings? 

$\bullet$ \textbf{RQ4.} What is the effect of our co-teaching mechanism from the qualitative perspective?

In what follows, we first setup the experiments, and then present and analyze the evaluation results to answer these questions.

\subsection{Experimental Setup}

\begin{table}[t]
	\centering
	\caption{{Statistics of the datasets. \#Accept/\#Reject  denote the number of explicit interaction that an employer sends notifications of acceptance or rejection;  \#Clicking denotes the number of interaction that an employer clicks the link to homepage of a job seeker and chats online after reviewing a resume, but does not send notifications; \#Browsing denotes the number of interaction that an employer only reviews a resume without subsequent behavior.}}
	\begin{tabular}{p{3.1cm}<{\centering}|p{1.4cm}<{\centering}|p{1.1cm}<{\centering}|p{1.1cm}<{\centering}}
		\hline\hline
		Statistics           &Technology   &Sales &Design  \\ \hline
		{\#Jobs}&21695&7964&7332\\\hline
		{\#Resumes}&35902&6731&10347\\\hline
		{\#Reject}&2875&567&1124\\\hline
		{\#Accept}&109125&16417&31293 \\\hline
		{Density}&0.0144\%&0.0317\%&0.0427\% \\\hline\hline
		{\#Clicking}&7468577&2523337&2615530 \\\hline
		{\#Browsing}&14110159&2484599&3863614\\\hline
		\hline
	\end{tabular}
	\label{tab-data}
\end{table}

\subsubsection{Datasets:}
We evaluate our model on a real-world dataset provided by a popular online recruiting platform named ``BOSS Zhipin'' (the BOSS Recruiting)\footnote{https://www.zhipin.com} in China. To protect the privacy of candidates, all the records have been anonymized by removing identity information.
The original dataset is split into three categories to test the robustness of our model for different domains. 
The statistics of the processed data are summarized in Table~\ref{tab-data}. We can see:
(1) All the datasets are extremely sparse, where the density ranges from 0.0144\% to 0.0427\%;
(2) Different categories correspond to varying data characteristics, \eg \emph{Technology} is a large and sparse dataset, while \emph{Sales} is much smaller but denser.
(3) For each category, the number of reject samples (\ie negative instances) is much smaller than that of the accept ones (\ie positive instances).
Since such an imbalanced dataset may bias the model learning process~\cite{he2009learning}, 
a commonly adopted method to balance the data distribution is to randomly sample job-resume pairs without explicit status~\cite{QinZXZJCX18}.
We consider two types of samples as negative instances: (1) an employer clicks the link to homepage of a job seeker and then chats online after reviewing her/his resume, but does not send notifications (\emph{clicking}), and (2) an employer reviews the resume without any further behavior (\emph{browsing}). As discussed before, sampled negative samples are likely to be ``\emph{false negative}'': although without explicit online acceptance notification, they may have actually reached the agreement on the job offer.  To incorporate such negative samples in training data, we would like to examine the capability of our model that learns from noisy data.
By equally sampling from these two resources, the final  ratio between the numbers of positive and negative samples is set to 1:1.
It should be noted that we only use these negative samples in the training stage, while for the validation and testing sets, we use the samples with explicit acceptance or rejection status to make sure our evaluation is accurate and reliable.

\begin{table*}[!t]
	\caption{{Performance comparison between the baselines and our model. 
			For each metric on different datasets, we use bold fonts and ``$^*$'' to mark the best performance and the best baseline performance, respectively.
			MV-CoN(F) and MV-CoN(R) are our models with instance filtering and re-weighting as the co-teaching strategies, respectively.
	}}
	\setlength{\tabcolsep}{5.pt}
	\begin{tabular}
		{p{1.6cm}<{\centering}
			|p{0.7cm}<{\centering}p{0.7cm}<{\centering}p{0.6cm}<{\centering}p{0.7cm}<{\centering}p{0.8cm}<{\centering}
			|p{0.7cm}<{\centering}p{0.7cm}<{\centering}p{0.6cm}<{\centering}p{0.7cm}<{\centering}p{0.8cm}<{\centering}
			|p{0.7cm}<{\centering}p{0.7cm}<{\centering}p{0.6cm}<{\centering}p{0.7cm}<{\centering}p{0.8cm}<{\centering}} \hline\hline
		
		Dataset & \multicolumn{5}{c|}{Technology} & \multicolumn{5}{c|}{Sales}& \multicolumn{5}{c}{Design}\\\hline
		
		Metric&AUC&ACC&P&R&$F_1$
		&AUC&ACC&P&R&$F_1$
		&AUC&ACC&P&R&$F_1$\\\hline
		
		DSSM    &0.690&0.636&0.645&0.636&0.640
		&0.682&0.629&0.640&0.631&0.635
		&0.696&0.641&0.649&0.638&0.644\\
		
		NFM   &0.652&0.608&0.613&0.602&0.607
		&0.654&0.601&0.612&0.606&0.609
		&0.668&0.618&0.627&0.620&0.624\\
		
		BPJFNN  &0.704&0.650&0.651&0.638&0.644
		&0.696&0.644&0.648&0.632&0.640
		&0.708$^*$&0.655&0.655&0.639&0.647\\
		PJFNN   &0.686&0.639&0.639&0.630&0.634
		&0.677&0.633&0.635&0.625&0.630
		&0.690&0.643&0.645&0.631&0.638\\
		APJFNN  &0.700&0.648&0.650&0.639&0.644
		&0.694&0.641&0.645&0.634&0.639
		&0.705&0.652&0.656&0.640&0.647\\
		
		JRMPM  &0.694&0.645&0.643&0.633&0.638
		&0.688&0.636&0.639&0.628&0.634
		&0.697&0.648&0.649&0.632&0.640\\
		
		DGMN    &0.699&0.651&0.652$^*$&0.639&0.645
		&0.693&0.643&0.648&0.635&0.642
		&0.704&0.655&0.656&0.641&0.648\\
		
		UBD &0.706$^*$&0.652$^*$&0.652$^*$&0.642$^*$&0.647$^*$
		&0.698$^*$&0.648$^*$&0.654$^*$&0.643$^*$&0.648$^*$
		&0.708$^*$&0.658$^*$&0.661$^*$&0.643$^*$&0.651$^*$\\
		
		\hline
		MV-CoN(F)  &0.728&0.672&0.679&0.646&0.664
		&0.725&$\textbf{0.671}$&0.677&0.640&0.659
		&0.736&0.679&0.682&0.646&0.663\\
		MV-CoN(R)  &$\textbf{0.737}$&$\textbf{0.678}$&$\textbf{0.683}$&$\textbf{0.649}$&$\textbf{0.668}$
		&$\textbf{0.727}$&0.669&$\textbf{0.679}$&$\textbf{0.644}$&$\textbf{0.661}$
		&$\textbf{0.743}$&$\textbf{0.683}$&$\textbf{0.688}$&$\textbf{0.651}$&$\textbf{0.669}$\\\hline\hline
	\end{tabular}\label{tab:main-results}
\end{table*}

\subsubsection{Baselines:}
We compare our model with the following representative methods:
\ignore{
\begin{itemize}
	\item \textbf{DSSM}~\cite{huang2013learning} leverages convolutional layers to extract semantic information for making the final prediction.\\
	\item \textbf{NFM}~\cite{0001C17} extends factorization machine (FM) with neural architectures to learn nonlinear and high-order interaction signals, where we use the textual and ID features as input.\\
	\item \textbf{BPJFNN}~\cite{QinZXZJCX18} leverages BiLSTM to derive the job and resume representations.\\
	\item \textbf{PJFNN}~\cite{ShenZZXMX18} is a CNN based method, and the matching degree is computed by the cosine similarity.\\
	\item \textbf{APJFNN}~\cite{QinZXZJCX18} proposes to use hierarchical recurrent neural networks to process the job and resume contents, and the final prediction is regarded as a classification problem.\\	
	\item \textbf{JRMPM}~\cite{YanLSZZ019} captures the preference information of the jobs and resumes by introducing a profiling memory module.\\
	\item \textbf{DGMN}~\cite{BianZSZW19} is a deep model focusing on the global sentence interactions when matching the jobs and resumes.\\
	\item \textbf{UBD}~\cite{MalachS17} is a method designed for learning from noisy labels, and updates parameters based on their disagreement between the two classifiers.\\
\end{itemize}
}

$\bullet$ \textbf{DSSM}~\cite{huang2013learning} leverages convolutional layers to extract semantic information for making the final prediction.

$\bullet$ \textbf{NFM}~\cite{0001C17} extends factorization machine (FM) with neural architectures to learn nonlinear and high-order interaction signals, where we use the textual and ID features as input.

$\bullet$ \textbf{BPJFNN}~\cite{QinZXZJCX18} leverages BiLSTM to derive the job and resume representations.

$\bullet$ \textbf{PJFNN}~\cite{ShenZZXMX18} is a CNN based method, and the matching degree is computed by the cosine similarity.

$\bullet$ \textbf{APJFNN}~\cite{QinZXZJCX18} proposes to use hierarchical recurrent neural networks to process the job and resume contents, and the final prediction is regarded as a classification problem.

$\bullet$ \textbf{JRMPM}~\cite{YanLSZZ019} captures the preference information of the jobs and resumes by introducing a profiling memory module.

$\bullet$ \textbf{DGMN}~\cite{BianZSZW19} is a deep model focusing on the global sentence interactions when matching the jobs and resumes.

$\bullet$ \textbf{UBD}~\cite{MalachS17} is a method designed for learning from noisy labels, and updates parameters based on their disagreement between the two classifiers.

Our baselines have covered most of the recently proposed job-resume matching models with different model architectures. 

\subsubsection{Evaluation and implementation details} 
Following previous work~\cite{QinZXZJCX18,ShenZZXMX18}, we adopt the evaluation metrics including \textbf{AUC},  \textbf{Accuracy}, \textbf{Precision}, \textbf{Recall} and \textbf{F1} to evaluate our models.

For each category, we first split the augmented dataset  into train, validation, and test sets with an approximate ratio of 8:1:1. Note that all the instances in validation and tests are 
guaranteed to be with ground-truth labels. While, the negative instances in training set are likely to contain noisy labels, as they are sampled from the interaction pairs without explicit status in the recruitment platform (see Section 5.1.1).  



When tuning our model parameters, the sentence embeddings are initialized via the BERT-Base-Chinese\footnote{https://github.com/huggingface/transformers}, and the document representation is derived as the output of the Transformer encoder.
The filter parameter $\delta$ and the number of Transformer layers are varied in the ranges of $\{0.2, 0.4, 0.6, 0.8, 1.0\}$ and $\{1,2,3,4\}$, respectively.
The batch size is empirically set to 32. 
The Adam optimizer is used to learn our model, and the learning rate is tuned in $\{0.01, 0.005, 0.001, 0.0005, 0.0001\}$.
Early stopping is used with a patience of 5 epochs.
For the baselines, the parameters are set as their default values or tuned on the validation set. Our dataset and code are available at this link: ~\changed{\url{https://github.com/RUCAIBox/Multi-View-Co-Teaching}}.

\subsection{RQ1: The Overall Comparison}
Table~\ref{tab:main-results} presents the comparison between our model and the baselines.
First, the feature interaction method NFM is difficult to achieve good results on our task. The main reason is that the sparsity level of the interaction between job and resume is much smaller than that of traditional recommendation systems. Meanwhile, DSSM does not perform well in most cases because it fails to capture the sequential properties in textual information.
The performance difference between BPJFNN, PJFNN, APJFNN, JRMPM and DGMN is small, and the winner varies on different metrics or datasets. Furthermore, UBD is the only baseline that specially learns from noisy labels, which updates parameters using the instances with different predictions from multiple classifiers. As we can see that, it substantially improves over the other baseline methods on all the datasets, indicating the necessarity of handling noisy labels.

As a comparison, our model achieves the best performance on all the metrics across different datasets. 
In specific, our model can on average improve the best baseline by 3.1\%, 2.9\% and 3.5\% on the datasets of Technology, Sales and Design, respectively.
This result positively answers our first research question (\textbf{RQ1}), and verifies the effectiveness of our designed multi-view co-teaching network.
Compared with the baselines, the co-teaching mechanism in our model is able to identify more informative and reliable samples for learning the parameters.
Our model is potentially more resistible to the negative impact brought by noisy data, and thus performs better than the other comparison methods.
Comparing the two strategies of instance filtering and re-weighting, we find that the latter is better in most cases. A possible reason is that re-weighting strategy adopts a ``soft'' denoising approach  that is more robust in dealing with noisy  labels.

\subsection{RQ2: Ablation Study}
To effectively learn from sparse, noisy interaction data for job-resume matching, our model has incorporated three technical components including text-based matching model (denoted by \underline{T}), relation-based matching model (denoted by \underline{R}) and co-teaching mechanism (denoted by \underline{C}). Here, we examine how each of them affects the final performance. 

\begin{table}[t]
	\centering
	\caption{Effects of different model components. Here, T denotes text-based component, R denotes relation-based component and C denotes co-teaching component.}
	\begin{tabular}{p{1.9cm}<{\centering}|p{.8cm}<{\centering}|p{.8cm}<{\centering}|p{.8cm}<{\centering}|p{.8cm}<{\centering}|p{.8cm}<{\centering}}
		\hline\hline
		Metric&AUC&ACC&P&R&$F_1$ \\ \hline
		R&0.706&0.651&0.653&0.641&0.647 \\
		T&0.717&0.663&0.665&0.653&0.659 \\
		TR&0.712&0.654&0.659&0.648&0.653 \\\hline
		TTC&0.720&0.666&0.668&\textbf{0.657}&0.662\\
		RRC&0.714&0.655&0.662&0.647&0.654\\
		TRC&\textbf{0.737}&\textbf{0.678}&\textbf{0.683}&0.649&\textbf{0.668}\\\hline\hline
	\end{tabular}
	\label{tab-ablation}
\end{table}

We consider the following five variants of our approach for comparison:
\textbf{(A)} \underline{R} is the single relation-based matching model, 
\textbf{(B)} \underline{T} is the single text-based matching model, 
\textbf{(C)} \underline{TR} is the simple  fusion model that directly averages the output from the text- and relation-based components, 
\textbf{(D)} \underline{TTC} is a co-teaching method that only utilizes text-based matching model, and
\textbf{(E)} \underline{RRC} is a co-teaching method that only utilizes relation-based matching model.
Here, \underline{TRC} denotes our proposed model. 
In the experiments, the parameters are remained as the default settings, taking re-weighting strategy for the co-teaching mechanism. 
Due to the space limit, we only report the results on the Technology dataset, and similar conclusions can be obtained on the other datasets.

In Table~\ref{tab-ablation}, we can see that the performance order can be summarized as: \underline{R} < \underline{TR} <  \underline{RRC} < \underline{T} < \underline{TTC} < \underline{TRC}. These results indicate that all the three components are indeed useful to improve the performance of job-resume matching. Especially, the text-based matching model and co-teaching mechanism bring more improvement to our approach. Besides, an interesting observation is that simply fusing the multi-view data may not lead to a good performance (\ie  \underline{TR} < \underline{T}).
It indicates that we need to carefully design the fusion strategy that utilizes multi-view data  for our task. 

\subsection{RQ3: Performance Tuning}
In this part, we examine the robustness of our model, and analyze the 
 influence of  parameters (or hyper-parameters) and training data on model performance. 
 For simplicity, we only incorporate the best baseline UBD from Table~\ref{tab:main-results} as a comparison.
 
\begin{figure}[H]
	\centering
	\subfigure[The  selection ratio of instance filtering.]{
		\label{fig:subfig:delta} 
		\includegraphics[width=1.60in]{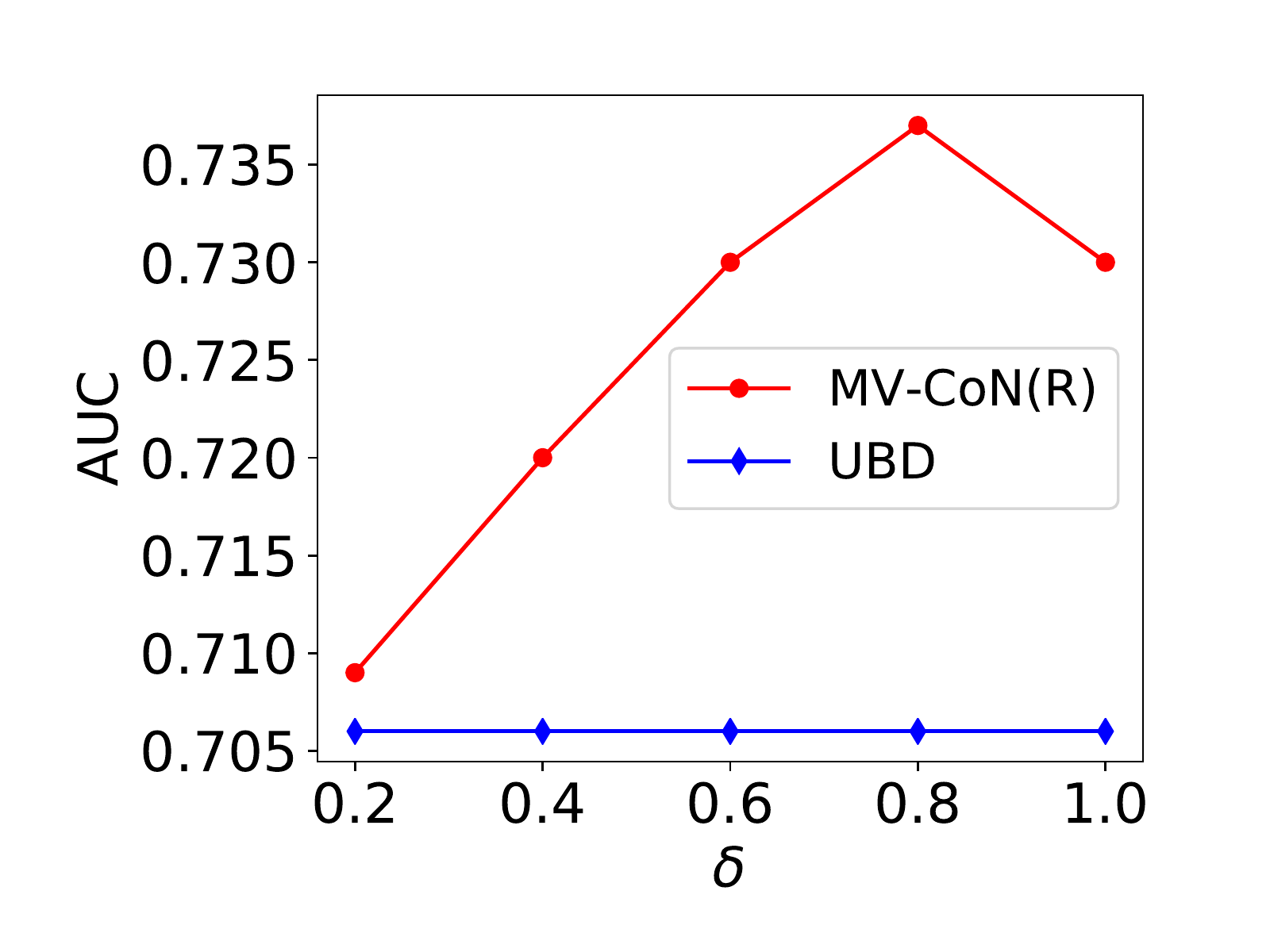}}
	\subfigure[The number of Transformer layers.]{
		\label{fig:subfig:l} 
		\includegraphics[width=1.60in]{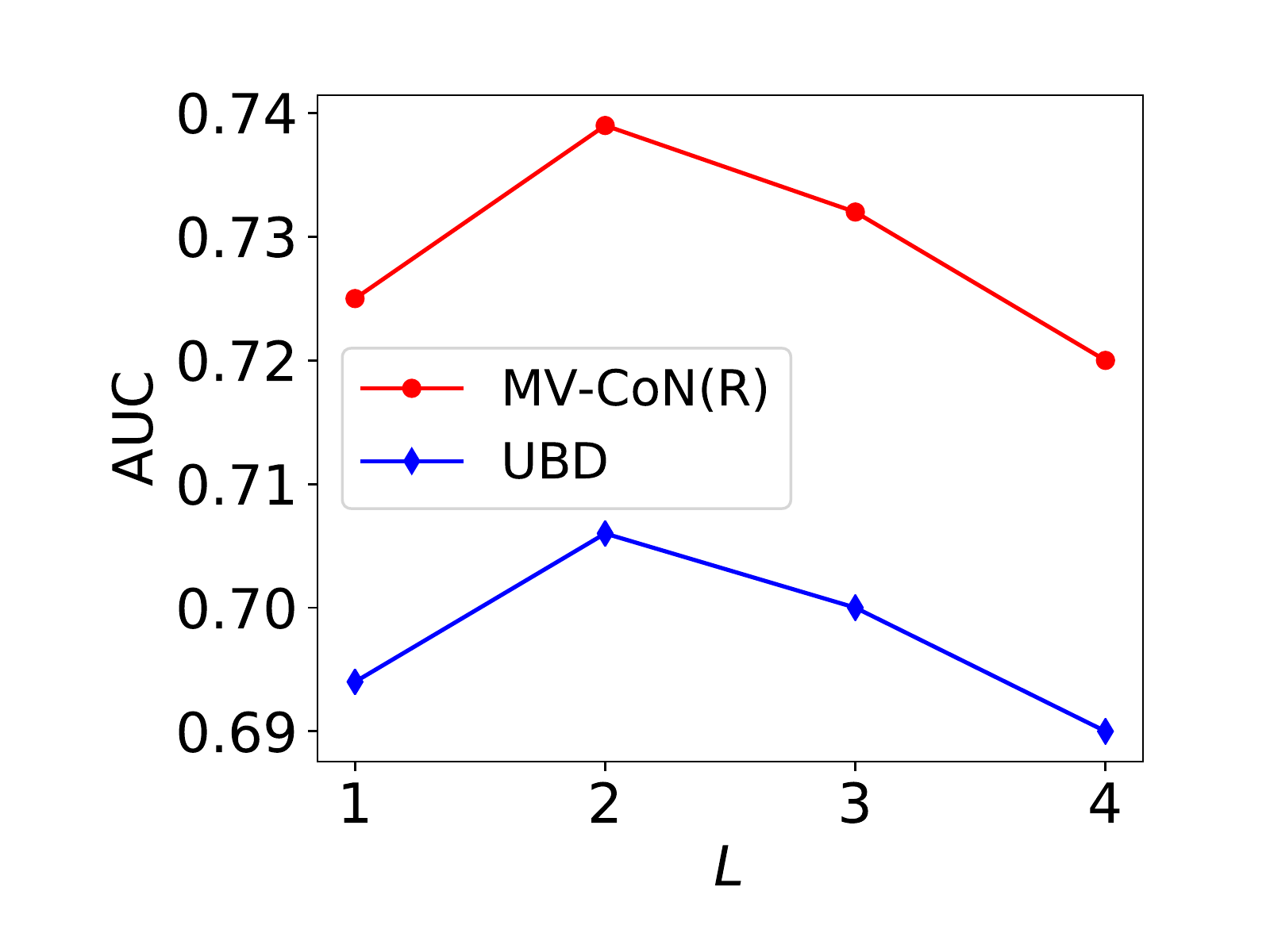}}
	\caption{Performance tuning with   the selection ratio of instance filtering strategy ($\delta$) and the number of Transformer layers ($L$).}
	\label{fig-parameters} 
\end{figure}

In our model, we have two important parameters, namely the selection ratio $\delta$ of instance filtering strategy and the number of Transformer layers $L$.
First, we vary the selection ratio $\delta$  in the set $\{0.2, 0.4, 0.6, 0.8, 1.0\}$. 
As shown in Eq.~\ref{eq-filter}, $\delta$ controls the data amount from a model for the peer model.  
 In Figure~\ref{fig:subfig:delta}, we can see that a selection ratio of $0.8$ achieves the best performance for our model.  As  $\delta$ decreases below 0.8, the performance continues to drop.   
It is because that when filtering more training instances, we also tend to exclude true instances from training data. 
When the training set  becomes small,  it is not sufficient to train a good peer model.  
Then, we vary the number of Transformer layers in the set $\{1, 2, 3, 4\}$. It can be observed from Figure~\ref{fig:subfig:l} that  two-layer Transformer achieves the best performance for our model.
While, the overall performance is relatively stable when we use more or fewer Transformer layers.

\begin{figure}[H]
	\centering
	\subfigure[Technology dataset.]{
		\label{fig:subfig:tech} 
		\includegraphics[width=1.60in]{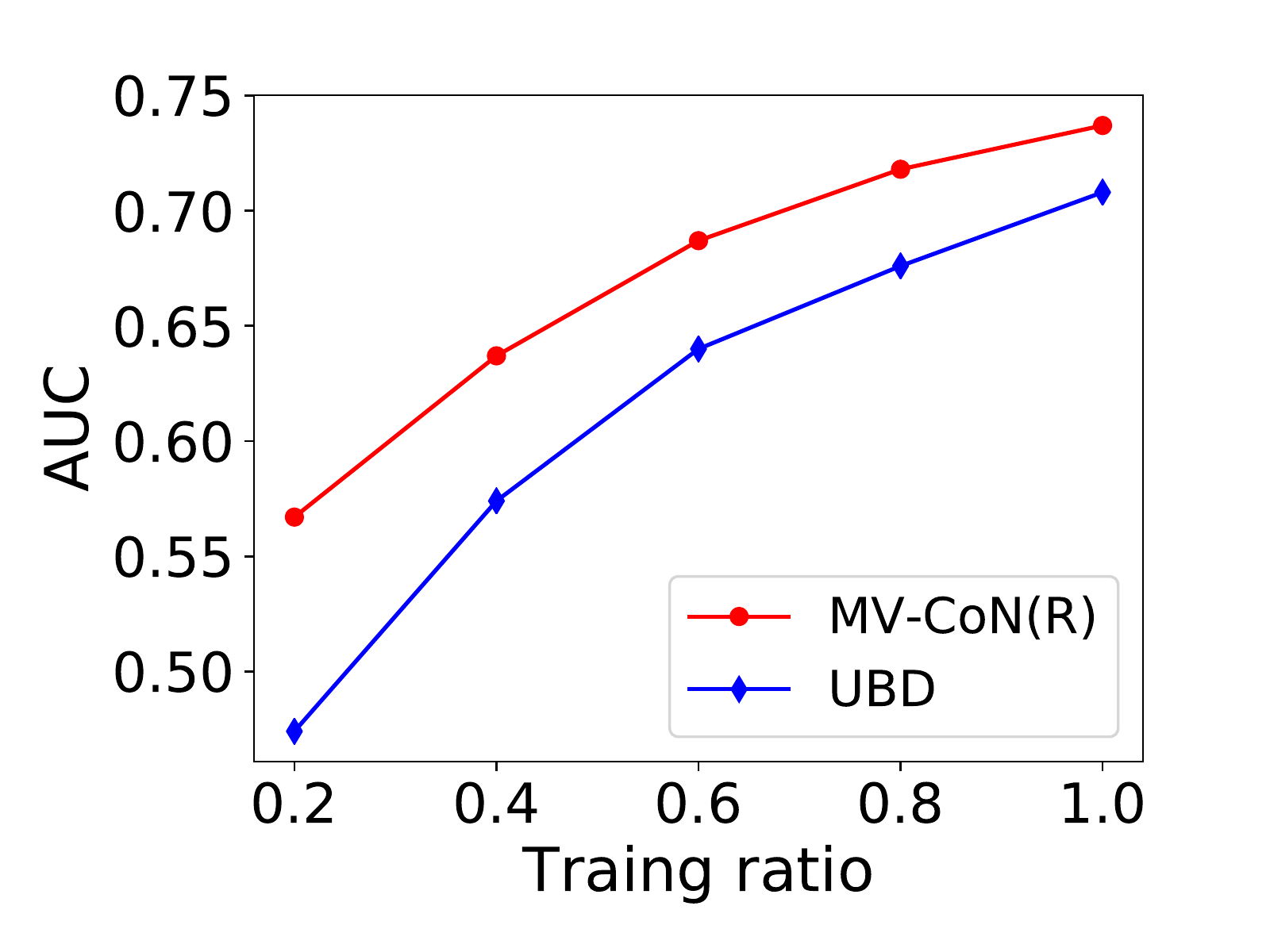}}
	\subfigure[Sales dataset.]{
		\label{fig:subfig:sales} 
		\includegraphics[width=1.60in]{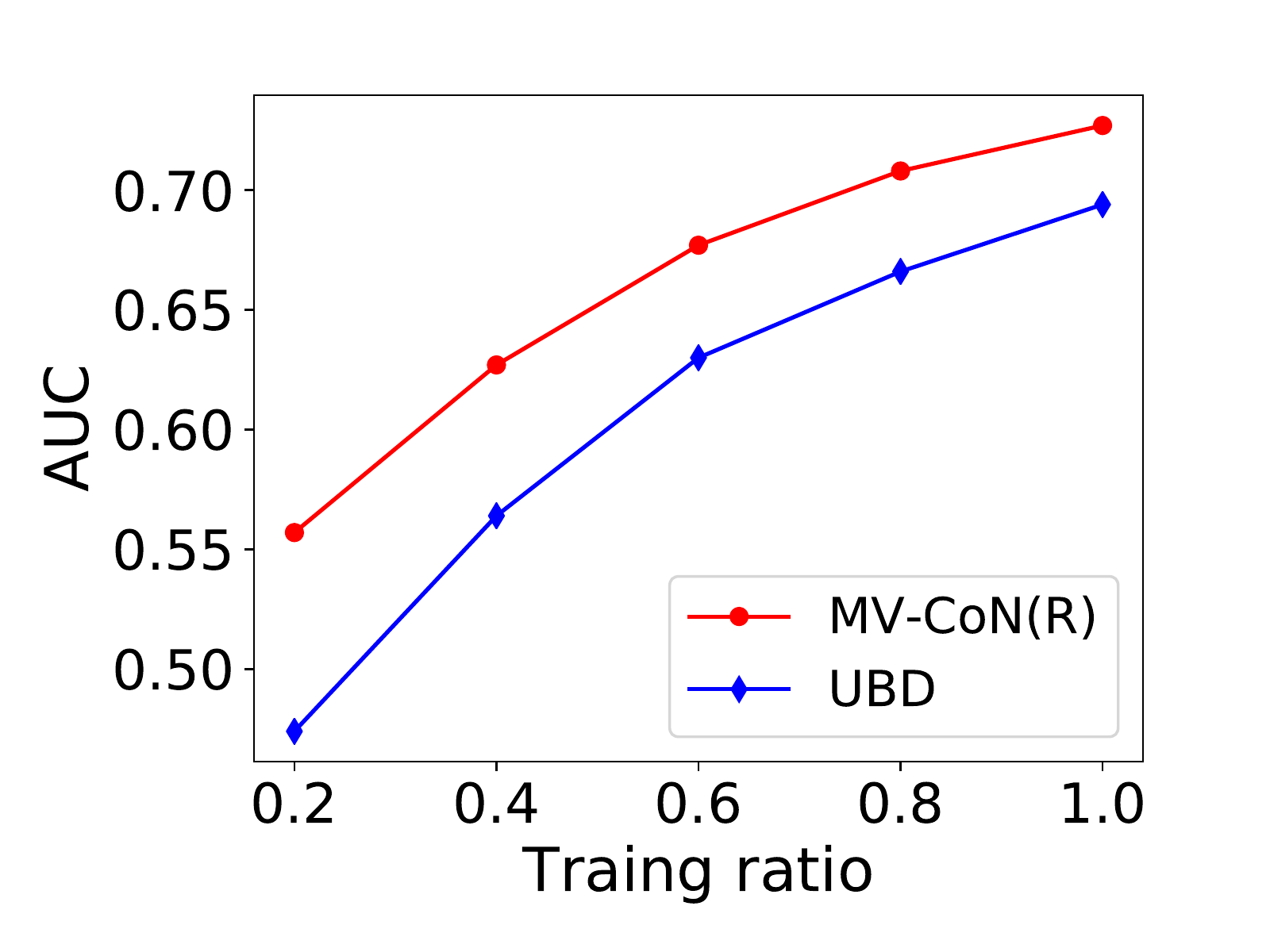}}
	\caption{Performance comparison by varying the amount of training data.}
	\label{fig-trainingdataamount} 
\end{figure}
Next, we study the influence of training data size on model performance.
\begin{figure*}[htbp]
	\centering
	\includegraphics[width=1.0\textwidth]{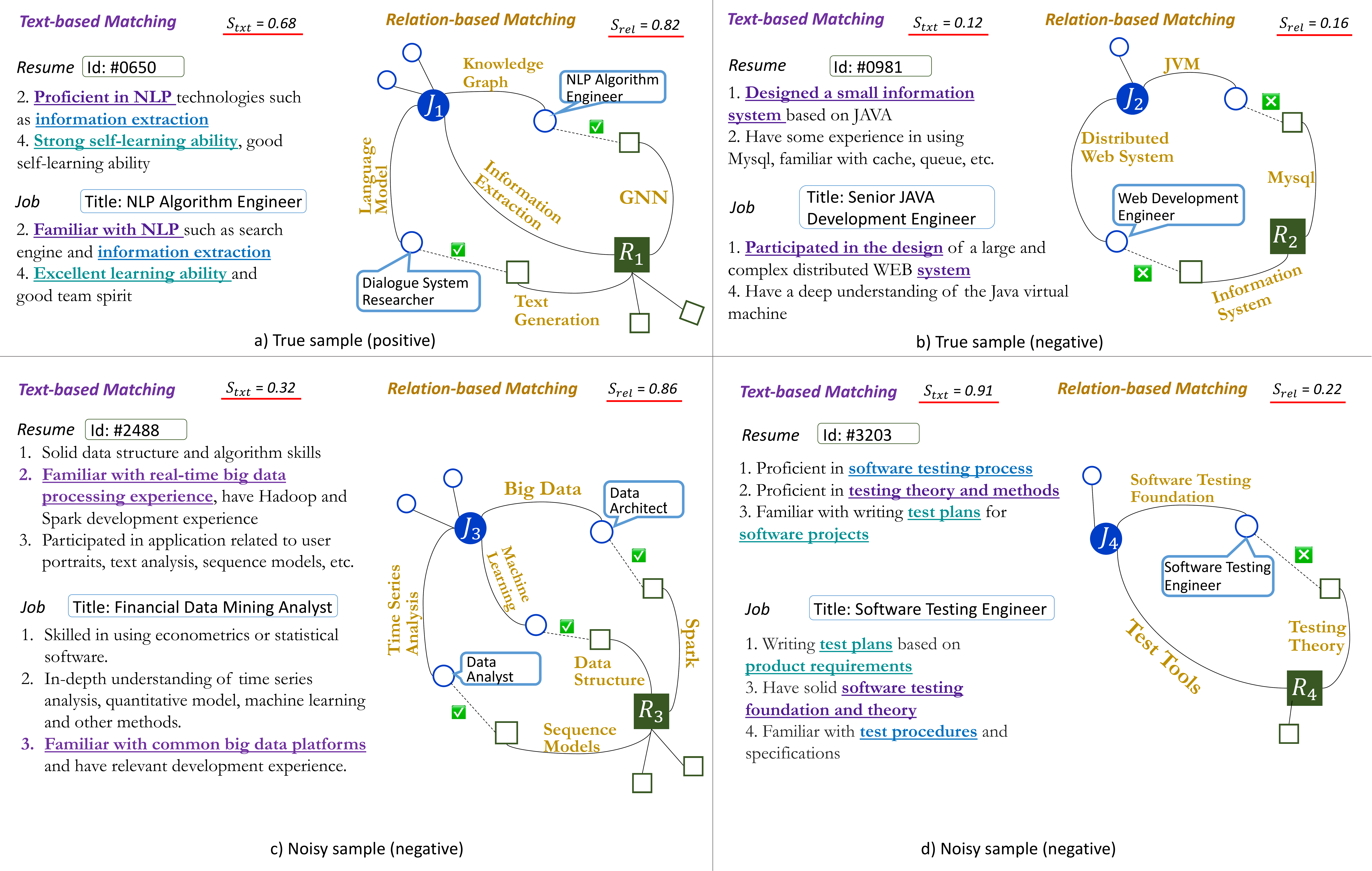}
	\caption{An illustrative example for showing the effect of different views in our co-teaching network. Here, we use circles and squares to denote jobs and resumes, respectively. For the links between two nodes, we attach the relation types or interaction labels. We mark the matched or similar text across jobs and resumes in underlined fonts. }
	\label{fig:case}
\end{figure*}
To examine this, we take 20\%, 40\%, 60\% and 80\% from the complete training data to generate four smaller training sets. We fix the test set as original, then learn the model with new training sets, and finally report the corresponding results on the test set. 
Figure~\ref{fig-trainingdataamount} presents the results of our model with different ratios of training data.

As we can see,  our model consistently outperforms the best baseline in Table~\ref{tab:main-results}. Especially, when training data is extremely sparse (\ie 20\%), the improvement  becomes more significant. These results show that our model is able to alleviate the influence of data sparsity to some extent, since we have utilized multi-view data in a principled way.

\ignore{, especially when training data is extremely sparse (\ie 20\%). 
which demonstrates the effectiveness of our model under sparse dataset settings.
This observation implies that our model is able to alleviate the influence of data sparsity to some extent. Besides, it can yield more improvement with less training data.
}

\subsection{RQ4: Qualitative Analysis}

Above, we mainly verify the effectiveness of our model in a quantitative manner.
It would be also helpful to present some case studies for understanding the working mechanism from a qualitative perspective.
In this part, we present four representative examples from the Technology dataset, and illustrate how our model works on them. 

In Figure ~\ref{fig:case}, the upper part corresponds to two true samples, where the acceptance or rejection status has been confirmed in the recruitment platform.
While, the bottom part corresponds to two noisy samples. They have been randomly sampled via 
via clicking or browsing interaction, originally labeled as negative samples in training data.
However, after the investigation by platform staff members, the two pairs turn out to be \emph{false negative}, \ie employment agreements have been reached offline for the two cases. 
Thus, the latter two samples are noisy instances, which are likely to affect the performance of the matching model.  

From cases (a) and (b), we can see, both of the text- and relation-based models produce consistent matching scores for the true samples, while their working mechanisms are quite different.
The text-based model assigns a high score (\emph{i.e.}, 0.68) to  case (a),  because many skills in the resume, such as \emph{``familiar with NLP''} and \emph{``excellent learning ability''}, have exactly satisfied the job requirements.
As a comparison, in the relation-based model, the matching result is mainly based on the relation-based connections (\emph{e.g.}, \emph{knowledge graph},   \emph{GNN} and \emph{information extraction}) and matching signals of neighbor nodes.
Similar results can be also observed from the negative instance of case (b). These examples show that, in our approach, different model views can usually obtain consistent matching results for the true samples, though their mechanisms are  different. 

From cases (c) and (d), we can see that the two models have made inconsistent predictions on the noisy negative samples. For example, in case (c),  since this position is a relatively new position (\ie \emph{Financial Data Mining Analyst}),  the text-based model assigns a low score to the candidate job-resume pair, while the relation-based model considers it as a positive instance due to the rich connections through the constructed relation graph. Similar to case (c), although the relation-based model in case (d) negates the instance, the text-based model assigns a high matching score since the job-resume documents are similar in contents. Once a negative sample receives inconsistent prediction scores, our model will consider them as low-quality negative samples and penalize them with re-weighting (Eq.~\ref{eq-weight}) or filtering  (Eq.~\ref{eq-filter}) strategy.  

These examples have shown the  capability of our model for noisy information filtering from a qualitative perspective.

\section{Conclusion}

This paper presented a multi-view co-teaching network that is able to learn from sparse, noisy interaction data for job-resume matching. 
We considered two views for developing the matching algorithm, namely text- and relation-based models. 
Furthermore, the two models were integrated into a unified approach that was able to combine their merits for this task. We designed two strategies for model integration, namely representation enhancement and data enhancement. 
Representation enhancement referred to the sharing of the learned parameters or representations across the two models; data enhancement referred to the process of filtering or re-weighting training instances according to their quality, which was implemented by the co-teaching algorithms.
Extensive experiments showed that the proposed approach is able to achieve a better matching performance from  sparse and noisy interaction data by comparing with several competitive baselines. 
 
 In this paper, we only focus on the macro interaction behaviors, \ie the acceptation of interview or rejection. While, it is intuitive that other kinds of micro interactive actions should be also useful to the matching task, such as \emph{click} or \emph{dwell time}. We will investigate into this topic and develop a more comprehensive interaction model. Besides, we will also consider applying our approach to more categories and study the domain adaptation problem across different categories. 
\section*{Acknowledgement}
This work was partially supported by the National Key R\&D Program of China (2019AAA0105200), the National Natural Science Foundation of China under Grant No. 61872369 and 61832017, Beijing Academy of Artificial Intelligence (BAAI) under Grant No. BAAI2020ZJ0301, and Beijing Outstanding Young Scientist Program under Grant No. BJJWZYJH012019100020098, the Fundamental Research Funds for the Central Universities, the Research Funds of Renmin University of China under Grant No.18XNLG22 and 19XNQ047. Xin Zhao is the corresponding author.

\bibliographystyle{ACM-Reference-Format}
\bibliography{cikm-base}

\end{document}